\documentclass[12pt]{iopart}
\usepackage{color}
\usepackage{iopams} 
\usepackage{subfigure}
\usepackage{epsfig}
\usepackage{graphicx}
\usepackage{epstopdf}
\expandafter\let\csname equation*\endcsname\relax
\expandafter\let\csname endequation*\endcsname\relax
\usepackage{amsmath,amssymb}    
\usepackage[numbers]{natbib}
\usepackage[titletoc,toc,title]{appendix}
\usepackage{graphicx}
\usepackage{subfig}
\usepackage{caption}
\usepackage{sidecap}
\usepackage[applemac]{inputenc}
\usepackage{natbib}
\usepackage{graphicx}           
\usepackage{flafter}            
\usepackage{memhfixc}           
\begin{document}

\title[ ~]{Analysis of Random Sequential Message Passing Algorithms for Approximate Inference}

\vspace{0.1pc}
\author{Burak \c{C}akmak$^1$, Yue M. Lu$^2$ and Manfred Opper$^{3}$}

\vspace{0.3pc}
\address{$^1$ Artificial Intelligence Group, Technische Universit\"{a}t Berlin, 10587, Germany}
\vspace{0.2pc}
\address{$^2$ John A. Paulson School of Engineering and Applied Sciences, Harvard University,  Cambridge, MA 02138, USA}
\vspace{0.2pc}
\address{$^3$ Centre for Systems Modelling and Quantitative Biomedicine, University of Birmingham, B15 2TT, United Kingdom }


\begin{abstract}
We analyze the dynamics of a random sequential message passing algorithm for approximate inference with large Gaussian latent variable models in a student-teacher scenario. To model nontrivial dependencies between the latent variables, we assume random covariance matrices drawn from rotation invariant ensembles. Moreover, we consider a model mismatching setting, where the teacher model and the one used by the student may be different. By means of dynamical functional approach, we obtain exact dynamical mean-field equations characterizing the dynamics of the inference algorithm. We also derive a range of model parameters for which the sequential algorithm does not converge. The boundary of this parameter range coincides with the \emph{de Almeida Thouless} (AT) stability condition of the \emph{replica symmetric} ansatz for the static probabilistic model.
\end{abstract}
%




\def\mathlette#1#2{{\mathchoice{\mbox{#1$\displaystyle #2$}}%
                               {\mbox{#1$\textstyle #2$}}%
                               {\mbox{#1$\scriptstyle #2$}}%
                               {\mbox{#1$\scriptscriptstyle #2$}}}}
\newcommand{\matr}[1]{\mathlette{\boldmath}{#1}}
\newcommand{\RR}{\mathbb{R}}
\newcommand{\CC}{\mathbb{C}}
\newcommand{\NN}{\mathbb{N}}
\newcommand{\ZZ}{\mathbb{Z}}
\newcommand{\at}[2][]{#1|_{#2}}
\def\oneh{\frac{1}{2}}
\newcommand{\Kb}{{\mathbf{K}}}
\section{Introduction}
The analysis of the dynamics of message passing algorithms for inference in large probabilistic models has attracted
considerable interest in the fields of statistical physics and information sciences \cite{Kabashima,Bolthausen,Bayati,Mimura,Opper16,CakmakOpper19,cakmak_2020,takeuchi,rangan2019vector,fletcher2018inference,fan2020approximate}. From a statistical physics point of view, 
the fixed points of such algorithms correspond to solutions of TAP mean field equations for disordered systems \cite{TAP,Parisi,Adatap,Kab08}. The latter, under some conditions on the statistics of the disorder, can lead to {\em exact solutions} to thermal averages in the large system limit. Hence, message passing algorithms provide efficient computation methods for obtaining accurate solutions to high--dimensional statistical inference problems.

So far, most of the theoretical works on the dynamics of message passing 
consider a parallel update scheme, where all dynamical nodes are updated {\em simultaneously} at each iteration of the algorithm. For large classes of the random interaction matrices, the exact temporal progress of the algorithm can then be described by the so-called {\em state-evolution} equations \cite{Kabashima,Bolthausen,Donoha}.

In many applications, the parallel dynamics of the algorithm is often replaced by a sequential version, where only a subset of nodes is updated per iteration. For example, Minka's EP (expectation propagation)  algorithm \cite{Minka1}, which is one of the motivations behind the so-called VAMP (vector approximate message passing) approach \cite{Ma,rangan2019vector}, is originally formulated in terms of sequential iterations. This type of sequential algorithms have lower computational complexity per iteration. They can also be more memory efficient as they only need to have access to a small batch of the available data at any given time. Moreover, in certain situations, they were found to improve the convergence properties \cite{manoel15}.  Our goal in this paper is to extend the theoretical analysis of message passing dynamics from the parallel update setting to the sequential setting. Specifically, we address the following issues:

\begin{enumerate}
\item We analyze the dynamics of a random sequential message passing algorithm for approximate inference with a large  Gaussian latent variable model. At each iteration, a random selection of nodes are updated by the algorithm. The probability for a given node to be included in an update is a free parameter. Varying this parameter allows for an interpolation between a full parallel update of all nodes and the case where on average only a single node is updated. Relying on the technique of the {\em dynamical functional} approach of statistical mechanics \cite{Martin}, we decoupled the degrees of freedom and derive an effective \emph{single node} evolution equation that characterizes the limiting dynamics of the sequential algorithm.

\item  In practice, the probabilistic model assumed by the inference algorithm may differ significantly from the real data generating process. We take into account this issue by allowing for a possible mismatch between the data generating teacher model and the model used by the student. From a technical point of view, this more general scenario requires a larger number of time dependent order parameters to describe the dynamics of the algorithm. In addition, unlike the case  of perfect match between the student and teacher models \cite{ccakmak2020analysis}, the message-passing algorithm is no longer guaranteed to converge in the mismatched case. We have identified a range of model parameters for which the convergence of the sequential algorithm is impossible. Interestingly, the boundary of this parameter range coincides with the \emph{de Almeida Thouless} (AT) stability condition of the \emph{replica symmetric} ansatz for the probabilistic model \cite{Adatap,Kab08}.
\end{enumerate}

There have been several earlier studies of sequential dynamics for solving various statistical physics and inference problems \cite{sommers1987path,sollich1997line,mignacco2020dynamical}. The effective single node dynamics obtained in these studies often contain memory terms 
that make it difficult to evaluate the two-time correlation functions. Remarkably, due to the construction of our message passing algorithm, its single node dynamics has no \emph{memory term}. As a result, the corresponding two--time correlation functions can be obtained by tractable recursion formulas. A similar ``memory-free'' property of sequential algorithms was observed in our previous paper \cite{Cakmak21} on solving the TAP equations for the Sherrington--Kirkpatrick model. Finally, the issue of data-model mismatch has also been previously considered in \cite{takahashi2020macroscopic} for parallel-updating message passing algorithms. Unlike in \cite{takahashi2020macroscopic} where the analysis is focused on the ``single-time'' statistics of the algorithm, we characterize the full effective single-node dynamics. This characterization provides information about the joint statistics of the algorithm over multiple time steps, which is crucial for analysing the convergence properties of the message passing algorithm.


The paper is organized as follows: Section~\ref{sec:model} presents the details of the Bayesian probabilistic model considered in this work. We introduce in Section \ref{sec:algorithm} a random sequential iterative algorithm for solving the inference problem. Its thermodynamic properties are studied in Section~\ref{sec:DFT} by using the method of dynamical functional theory. Comparisons
of the theory with simulations are given in Section~\ref{Sim}. We conclude the paper in Section~\ref{sec:conclusion} with a summary and
some discussions. The derivations of our results can be found in the Appendix.

\section{Latent Gaussian variable models}
\label{sec:model}

Message passing algorithms have been successfully applied to latent Gaussian variable models \cite{Minka1,OW0,Rasmussen}. This class of models finds widespread applications
in statistics, machine learning and signal processing. A typical scenario is to infer an unobserved latent vector $\matr \theta\in\Re^{N\times 1}$ by using the Bayesian {\em posterior distribution} 
\begin{equation}
p(\matr \theta \vert\matr y,\matr K)\doteq\frac 1 Z \mathcal N(\matr \theta \vert \matr 0,\matr K)\prod_{i\leq N}p(y_i \vert \theta_i)\label{Gausslate}
\end{equation}
where $Z$ is a normalization constant. This model assumes that the components 
of the vector  $\matr y$ 
of $N$ real data values  are assumed to be generated independently from a likelihood  $p(y \vert \theta)$
based on a vector of unknown parameters $\matr \theta$. Prior statistical knowledge about 
$\matr \theta$ is introduced by the correlated Gaussian with covariance $\matr K\in \Re^{N\times N}$. 

We will later illustrate our theory on the well known example of Bayesian learning 
of a noisy perceptron---also known as {\em probit regression} \cite{neal1997monte}. This 
corresponds to a binary classification problem with class labels $y_i = \pm 1$. For this model, one assumes 
a training set given by $\{(\matr x_i ,y_i)\}_{i\leq N}$ where $\matr x_i\in \RR^{P\times 1}$ stands for a vector of inputs. Class labels $y_i$ 
are generated according to the observation model
\begin{equation}
y_i=\epsilon_i\,{\rm sign}(\matr x_{i}^\top\matr w+n_i).
\end{equation}
Here, we allow for additive i.i.d. Gaussian noises $n_i$ for all $i$ with $\matr n\sim \mathcal N(\matr 0,\sigma^2{\bf I})$ 
as well as i.i.d. multiplicative flip noises $\epsilon _{i}=\pm1$ for all $i$ with $\beta\doteq {\rm Pr}(\epsilon_i=-1)$. 
We assume a latent vector $\matr w$ with Gaussian prior distribution $\matr w\sim \mathcal N(\matr 0,{\bf I})$. To map this problem onto the model
\eqref{Gausslate}, we introduce the latent vector $\matr \theta=\matr X\matr \omega$ with $\matr X\doteq [\matr x_1^\top,\matr x_2^\top,\cdots, \matr x_N^\top]$. Hence, the prior covariance of $\matr \theta$ equals $\matr K=\matr X\matr X^\top$ and we have the data likelihood function
\begin{equation}
p(y\vert\theta)=(1-\beta)\Phi\left(\frac{y\theta}{\sigma}\right)+\beta \Phi\left(-\frac{y\theta}{\sigma}\right)\label{PercLike}
\end{equation} 
where $\Phi(\cdot)$ denotes the cumulative distribution function of the standard normal distribution.  

\section{The random sequential VAMP algorithm}
\label{sec:algorithm}

Typical prediction tasks based on observed data involve the computations of {\em expectations}
of components of $\matr \theta$ (or of functions of these components) using the posterior \eqref{Gausslate}. Unfortunately, except for
simple Gaussian likelihoods or simple diagonal covariance matrices, such expectations lead to multi--dimensional  integrals which cannot be computed analytically. Hence,  one has to resort to 
approximations. To be able to obtain reliable results in the case of high--dimensional vectors $\matr \theta$, so--called message passing algorithms have been developed which provide efficient iterative computations of generalized mean field approximations to the desired expectations.


Given the auxiliary single-site partition function 
\begin{align}
Z_\nu(\gamma,y)\doteq \int {\rm d }\theta\;p(y\vert \theta) e^{-\frac{\nu}{2}\theta^2+\gamma\theta}
\end{align}
the logarithmic derivatives 
\begin{align}
m_\nu(\gamma,y)&\doteq \frac{{\partial}\ln{Z_\nu(\gamma,y)}}{{\partial}\gamma} \approx \mathbb E[\theta \vert\matr y,\matr K] \label{func1}\\
m_\nu'(\gamma,y)&\doteq \frac{{\partial }m_\nu(\gamma,y)}{{\partial}\gamma} \approx \mathbb E[\theta^2 \vert\matr y,\matr K] - \mathbb E^2[\theta \vert\matr y,\matr K]\label{func2}
\end{align}
provide approximations to posterior mean and variances of single components $\theta \equiv \theta_i$
upon convergence of the algorithm. The mean $\gamma_i$ of the {\em cavity fields} of a node \cite{Adatap,CakmakOpper18} can be computed iteratively by the VAMP algorithm \cite{Ma,rangan2019vector}. In the following, we introduce a {\em random sequential} version of the usual parallel VAMP.  Before the iteration starts, we compute the spectral decomposition
\begin{equation}
\matr K=\matr O\matr D\matr O^{\top}
\end{equation}
where $\matr D$ is diagonal with the diagonal entries being the eigenvalues of $\matr K$. 
We define the iterative algorithm in discrete time by the vector updates  for
$t=1,2,\ldots T$ by
\begin{subequations}
	\label{svamp1}
\begin{align}
\matr \gamma^{(t)}&= \matr \gamma^{(t-1)} +\matr P^{(t)}[\matr \phi^{(t)}-\matr \gamma^{(t-1)}] \label{svamp1a}\\
\matr \phi^{(t)}&= \frac {1} {\tau^{(t)}}\matr O     \matr D(\lambda^{(t)}\matr D+{\bf I})^{-1}          \matr O^\top \matr {\tilde \gamma}^{(t)}-\tilde{\matr \gamma}^{(t)}\label{phit} \\
\matr {\tilde \gamma}^{(t)}&=\frac{m_{\nu^{(t-1)}}(\matr \gamma^{(t-1)},\matr y)}{\chi^{(t)}}-\matr\gamma^{(t-1)}.
\end{align}
\end{subequations}
The scalar quantities $\chi^{(t)}$, $\lambda^{(t)}$, $\tau^{(t)}$, and $\nu^{(t)}$  are updated as
\begin{subequations}
	\label{svamp2}
\begin{align}
\chi^{(t)}&=\langle m_{\nu^{(t-1)}}'(\matr \gamma^{(t-1)},\matr y)\rangle\\
\lambda^{(t)}&=\frac{1}{\chi^{(t)}}-\nu^{(t-1)}\\
\tau^{(t)}&=\frac{1}{N}\sum_{i\leq N} \frac{D_{ii}}{\lambda^{(t)} D_{ii}+1}\\
\nu^{(t)}&=\frac{1}{\tau^{(t)}}-\lambda^{(t)}
\end{align}
\end{subequations}
where the brackets $\langle\ldots\rangle$ denote an {\em empirical average} over the sites. Moreover, we consider a random initialization for $\matr\gamma^{(0)}$ from an i.i.d. normal Gaussian distribution. The diagonal matrix $\matr P^{(t)}$ in \eqref{svamp1a} is composed of binary decision variables
$p_i^{(t)}\doteq P^{(t)}_{ii}\in\{0,1\}$. The original parallel version of the VAMP algorithm is obtained when 
$\matr P^{(t)}$ is equal to the unit matrix. Random {\em sequential} updates are introduced by 
making the $p_i^{(t)}$ random variables which decide
if node $i$ is updated ($p_i^{(t)} =1$) at time $k$ or not  ($p_i^{(t)} =0$).
We assume that the $p_i^{(t)}$ are independent for all $i,t$  and that
$ {\rm  Pr}(p_i^{(t)}=1) = \eta$. The case $\eta = 1/N$ corresponds to updating only a single node on average. 

\section{The dynamical mean-field equations}
\label{sec:DFT}

We consider an average case analysis of the algorithm in the limit $N\to\infty$ assuming that the data $\matr y$ is generated from a 
given likelihood model, where for generality, we consider a data-model mismatching scenario. 
In general, we assume 
that the components of the vector $\matr y$ are generated independently from a likelihood $p_0(y \vert \theta)$ which 
is not necessarily equal to $p(y\vert \theta)$.
For the noisy perception model this would correspond to different sets of hyperparameters $\beta$ and $\sigma$ in  
\eqref{PercLike}. But we also assume that the prior covariance matrix $\matr K$ is the same for both true parameter and the parameter in 
the inference (student) model. We choose $\matr K$ to be a 
random matrix with a rotational invariant distribution. This 
means that the matrix $\matr O$ in \eqref{svamp1} is assumed to be a {\em Haar} matrix, i.e. a random rotation.
In this way, it is possible to model matrices $\matr K$ with nontrivial (weak) dependencies between entries. 

Following previous studies \cite{Opper16,ccakmak2020analysis}, we derive an effective dynamics of a {\em single node}. This is obtained by averaging the generating functional
of the dynamics over the randomness of $\matr y$, $\matr O$ and $\{\matr P^{(t)}\}$ and a subsequent decoupling of the degrees of freedom. This involves
order parameter functions which are self-averaging for $N\to\infty$. Generating functionals are partition functions for
the computation of expectations of dynamical variables where the dynamics is included in terms of Dirac 
$\delta$ functions. To avoid cluttered notation, and with $\matr J \doteq \matr K^{-1}$ \footnote{ Unless the covariance matrix $\matr K$ has an inverse, one can consider the substitution $\matr K\to \matr K+\epsilon {\bf I}$ for $\epsilon>0$ and perform the limit $\epsilon \to 0$ at the end of the analysis. In any case, the need for the inverse $\matr K^{-1}$ will be bypassed in the analysis. Hence, without loss of generality, we can assume that $\matr K$ has an inverse.}, the dynamical functionals corresponding to \eqref{svamp1} and \eqref{svamp2} for $T$ discrete time steps
can be written in the form
\begin{align}
Z_i(\{l^{(t)}\})=&\int \prod_{t=1}^{T}\left\{ {\rm d}\matr \psi^{(t)}{\rm d}\matr m^{(t)}\;
\delta \left[\matr m^{(t)}- f_t\left(\{\matr\psi^{(t)}, \matr m^{(l)}, \matr P^{(l)}\}_{l=1}^{t};\matr y\right)\right] \nonumber \right.  \\
& \left.  \qquad \qquad\qquad \qquad  \times \delta(\matr \psi^{(t)}-\matr J \matr m^{(t)}){\rm e}^{i \psi_i^{(t)}l^{(t)}}\right\}
\end{align}
where $\{f_t\}$ is an appropriate sequence of non-linear scalar functions. 
Using the Fourier representation of the Dirac measures
the averaged generating functional  is of the form
\begin{align}
&\mathbb E[Z_i(\{l^{(t)}\})]=\int {\rm d}\matr\theta{\rm d}\matr y{\rm dP}(\matr O)\; p_0(\matr y\vert \matr \theta)\mathcal N(\matr\theta\vert\matr 0,\matr K) \prod_{t\leq T}{\rm dP}(\matr P^{(t)})\;Z_i(\{l^{(t)}\})\\
&=c\int {\rm d}\matr\theta{\rm d}\matr y\;p_0(\matr y\vert  \matr \theta)\prod_{t\leq T}{\rm dP}(\matr P^{(t)}) {\rm d}\matr \psi^{(t)}{\rm d}\matr m^{(t)}{\rm d}\matr {\hat \psi}^{(t)} 
\delta \left[\matr m^{(t)}- f_t\left(\{\matr\psi^{(l)}, \matr m^{(l)}, \matr P^{(l)}\}_{l=1}^{t};\matr y\right)\right]\nonumber \\
&\quad  \times {\rm e}^{{\rm i}\sum_{k}(\matr{\hat \psi}^{(t)})^\top \matr \psi^{(t)}}{\rm e}^{{\rm i}\sum_{t}\psi_{i}^{(t)}l^{(t)}} \mathbb E_\matr O \left[{\rm e}^{-\frac{1}{2}\matr \theta^\top \matr J\matr \theta-{\rm i}\sum_{t\leq T}(\matr{\hat\psi}^{(t)})^\top\matr J\matr\psi^{(t)}}\right]\label{disorder}
\end{align}
where ${\rm dP}(\matr O)$ stands for the Haar invariant measure of the orthogonal group $O(N)$ and $c$ stands for a 
nonrandom term to ensure the normalization property $\mathbb E[Z(\{l^{(t)}=0\})]=1$.

Appendix A gives a short summary of details and references needed for the computations of the expectations 
and the subsequent decoupling of the degrees of freedom.  We find that the effective statistics of an arbitrary single node $\gamma^{(t)}\equiv \gamma_i^{(t)}$ (with similar definitions for other variables) of the algorithm \eqref{svamp1} and \eqref{svamp2}
 is given by the stochastic process 
\begin{subequations}
	\label{esp}
\begin{align}
\left(\theta,y,\phi^{(1:T)}\right)&\sim \mathcal N(\theta\vert 0, q)p_0(y\vert \theta)\mathcal N(\phi^{(1:T)}\vert \theta \mathcal {\hat B},\mathcal C)\\
\gamma^{(t)}&=\gamma^{(t-1)}+p^{(t)}(\phi^{(t)}+\gamma^{(t-1)})
\end{align}
\end{subequations}
where for short $\phi^{(1:T)}\doteq(\phi^{(1)}, \phi^{(2)},\ldots, \phi^{(T)})$ and $q\doteq \lim_{N\to \infty}\frac 1 N {\rm tr}(\matr K)$. 

Luckily, similar to previous results \cite{ccakmak2020analysis} obtained for the simpler scenario of parallel dynamics and matching teacher--student models, 
the effective dynamics does {\em not contain memory terms}. These terms are often encountered for the stochastic dynamics of
disordered systems \cite{sollich1997line,mignacco2020dynamical} and would render the driving process $\phi^{(1:T)}$ non Gaussian. This would preclude the computation of explicit analytical results for averages at finite time $t$ and one would have to resort to Monte--Carlo simulations \cite{Eisfeller} of the effective process \eqref{esp}.

The entries of the $T\times 1$ vector $\mathcal {\hat B}$ and the $T\times T$ covariance matrix $\mathcal C$ are recursively computed according to
\begin{align}
\mathcal {\hat B}^{(t)}&=\frac{\tau^{(t)}\zeta^{(t)}\mathbb E[\theta{\tilde \gamma}^{(t)}]}{1-q\tau^{(t)}\zeta^{(t)}}\label{Bhat}\\
\mathcal C^{(t,t')}&=\frac{\mathcal D^{(t,t')}+\mathcal Q^{(t,t')}\left(q\mathcal {\hat B}^{(t)} \mathcal {\hat B}^{(t')} +\mathcal {\hat B}^{(t)}\mathbb E[\theta\tilde\gamma^{(t')}]+\mathcal {\hat B}^{(t')}\mathbb E[\theta\tilde\gamma^{(t)}]+\mathbb E[\tilde\gamma^{(t)}\tilde\gamma^{(t')}]\right)}{1-\mathcal Q^{(t,t')}}\label{covphi}
\end{align}	
where we have introduced  the auxiliary dynamical order parameters 
\begin{subequations}
	\begin{align}
	\zeta^{(t)}&=\frac{{\rm R}(-\tau^{(t)})-1/q}{q-\tau^{(t)}}\\
	{\mathcal Q}^{(t,t')}&=\tau^{(t)}\tau^{(t')}\left\{\begin{array}{cc}
	\frac{{\rm R}(-\tau^{(t')})-{\rm R}(-\tau^{(t)})}{\tau^{(t)}-\tau^{(t')}}&t\neq t'\\
	{\rm R}'(-\tau^{(t)}) & \text{else}
	\end{array}\right.\\
	\mathcal{D}^{(t,t')}&=\frac{\hat {\mathcal B}^{(t)}\hat {\mathcal B}^{(t')}}{\zeta^{(t)}\zeta^{(t')}}\left\{\begin{array}{cc}
	\frac{\zeta^{(t)}-\zeta^{(t')}}{\tau^{(t)}-\tau^{(t')}}& t\neq t'\\
	\frac{\zeta^{(t)}-{\rm R}'(-\tau^{(t)})}{q-\tau^{(t)}}& \text{else}.
	\end{array}\right.
	\end{align}
\end{subequations} 
Here, the function $\rm R(\omega)$ is defined as
\begin{equation}
{\rm R }(\omega)\doteq {\rm G}^{-1}(\omega )-\frac 1 \omega\label{Rtrs}
\end{equation}
where ${\rm G}^{-1}$ denotes the functional inverse (w.r.t. decomposition) of the function
\begin{equation}
{\rm G}(z)\doteq \lim_{N\to \infty}\frac 1 N{\rm tr}(\matr K(z\matr K-\bf I)^{-1}).
\end{equation}
Note that, when $\matr K$ has an inverse, the function ${\rm  R}$ stands for the R-transform \cite{oxfordSpeicher} of the limiting spectral distribution of  $\matr K^{-1}$. In general, ${\rm R}(\omega)$  is well-defined and it is related to the limiting distribution of the non-zero eigenvalues of $\matr K$. Finally, the random field $\tilde{\gamma}^{(t)}$ stands for the effective stochastic process of an arbitrary component of $\matr {\tilde  \gamma}^{(t)}$ in \eqref{svamp1}. Specifically, we have
\begin{equation}
\tilde{\gamma}^{(t)}=\frac{m_{\nu^{(t-1)}}(\gamma^{(t-1)},y)}{\chi^{(t)}}-\gamma^{(t-1)}
\label{tildgam}
\end{equation}
where for convenience we have replaced the empirical averages in representing the dynamical order parameter in the algorithm, such as $\chi^{(t)}$, $\lambda^{(t)}$ and etc, by the averages w.r.t. the effective stochastic process \eqref{esp}, e.g. $\chi^{(t)}=\mathbb E[m'_{\nu^{(t-1)}}(\gamma^{(t-1)},y)]$.
We can see that the explicit computations of order parameter functions
require expectations of nonlinear functions of {\em pairs} of correlated Gaussian random variables. These can be performed easily by numerical quadrature.
To obtain a recursion for such order parameters, we note that the first line of \eqref{esp} implies
\begin{equation}
\mathcal C_\phi^{(t,t')}=\mathcal C^{(t,t')}+q\hat{\mathcal B}^{(t)}\hat{\mathcal B}^{(t')}
\label{Cphi2time}
\end{equation}
where $\mathcal C_\phi^{(t,t')}\doteq \mathbb E[\phi^{(t)}\phi^{(t')}]$.
Finally, the covariance of the $\gamma^{(t)}$ variables can be obtained from the second line of \eqref{esp} by averaging over the decision variables $p^{(t)}$
\begin{align}
&\mathcal C_\gamma^{(t,t')}=(1-\eta)^2\mathcal C_\gamma^{(t-1,t'-1)}+\eta^2\left[\mathcal C_\phi^{(t,t')}\right.\nonumber\\
&\left.+\sum_{l'=1}^{t'-1}(1-\eta)^{t'-l'}\mathcal C_\phi^{(t,l')}+\sum_{l=1}^{t-1}(1-\eta)^{t-l}\mathcal C_\phi^{(t',l)}\right]\;\label{covgamma}.
\end{align}
Combined with \eqref{covphi} and \eqref{tildgam}, we obtain a closed set of equations for the iterative 
computation of two time correlation functions.
We give explicit results of such computations together with comparisons to simulations of the algorithm 
for the perceptron model \eqref {PercLike} in section 5. In the following section, we will analyse the local convergence properties of the algorithm based on a recursion for 
the necessary {\em single time} order parameters.
\subsection{The fixed point solution}
We assume in the following that parameters of the probabilistic model and initial conditions 
are chosen in such a way that asymptotically for large times, the algorithm will converge to a fixed point.
We will then analyze the consistency of this assumption and establish a necessary criterion for convergence and show its relation to the AT line of the static learning model. Translated to the case of the single node dynamics of the dynamical mean field approach, we assume that 
$\gamma^{(t)}$ converges to a {\em static random variable} $\gamma^\star$ for $t\to \infty$.
It follows from \eqref{esp}, that $\gamma^\star = \phi^\star$ which is the limit of the Gaussian random variable $\phi^{(t)}$
which drives the dynamics. Specifically, we have
\begin{equation}
\left(\theta,y,\gamma^\star\right)\sim \mathcal N(\theta\vert 0, q)p_0(y\vert \theta)\mathcal N(\gamma^\star \vert \theta \mathcal {\hat B}^\star,\mathcal C^\star)
\end{equation} 
where $\mathcal {\hat B}^\star$ stands for the stationary solution of $\hat{\mathcal B}^{(t)}$ etc. It is easy to see that 
\begin{align}
\hat{\mathcal B}^\star=\zeta^\star\mathbb E[\theta m_{\nu^\star}(\gamma^\star,y)] \label{gen1}.
\end{align}
Then, it follows from \eqref{covphi} that 
\begin{equation}
\mathcal C^\star=\frac{(\hat{\mathcal B}^\star)^2}{\zeta^\star(q-\chi^\star)} +\left(\mathbb E[m_{\nu^\star} (\gamma^\star,y)^2]-\frac{(\hat{\mathcal B}^\star)^2}{(\zeta^\star)^2(q-\chi^\star)}\right){\rm R}'(-\chi^\star)\label{general}.
\end{equation} 
For example, in the teacher--student matching case, i.e. when $p_0(y\vert \theta)=p(y\vert\theta)$, we have 
\begin{equation}
\mathbb E[m_{\nu^\star}(\gamma^\star,y)^2]=\mathbb E[\theta m_{\nu^\star}(\gamma^\star,y)]=q-\chi^\star
\end{equation}
and the general solutions \eqref{gen1} and \eqref{general} simplify to 
$\mathcal{\hat B}^\star=\mathcal C^{\star}={\rm R}(-\chi^\star )-1/q $.
which agrees with our previous results \cite{ccakmak2020analysis}.
\subsection{Single-time recursion of dynamics and the AT instability criteria }
In order to study the convergence towards the fixed point over time, we need the covariances between the random variables
$\gamma^{(t)}$, $\phi^{(t)}$ and their asymptotic limits.
This can be obtained from recursions of the single time order parameters defined from the limits  $\mathcal C_{\phi,\gamma}^{(t)}\doteq \lim_{t'\to \infty}\mathcal C_{\gamma,\phi}^{(t,t')}$ (assuming the limits exist). 
Using \eqref{Cphi2time}, \eqref{covgamma} and \eqref{covphi}, we get
\begin{align}
\mathcal C_\gamma^{(t)}&=(1-\eta)\mathcal C_\gamma^{(t-1)}+\eta\mathcal C_\phi^{(t)}\label{Cgamma}\\
\mathcal  C_\phi^{(t)}&=\mathcal C^{(t)}+q\hat{\mathcal B}^{(t)}\hat{\mathcal B}^{\star}\label{Cphi1}\\
\mathcal C^{(t)}&=\frac{\mathcal D^{(t)}+\mathcal Q^{(t)}\left(q\mathcal {\hat B}^{(t)} \mathcal {\hat B}^{\star} +\mathcal {\hat B}^{(t)}\mathbb E[\theta\tilde\gamma^{\star}]+\mathcal {\hat B}^{\star}\mathbb E[\theta\tilde\gamma^{(t)}]+\mathbb E[\tilde\gamma^{(t)}\tilde\gamma^{\star}]\right)}{1-\mathcal Q^{(t)}}\label{Cphi2}
\end{align}
where e.g. $\mathcal Q^{(t)}\doteq \lim_{t'\to \infty}\mathcal Q^{(t,t')}$. In a similar way, we can show that
\begin{align}
\mathbb E[\theta\tilde\gamma^{(t)}]&=(1-\eta)\mathbb E[\theta\tilde\gamma^{(t-1)}]+\eta \mathbb E[\theta f_t(\phi^{(t-1)})]\\
\mathbb E[\tilde\gamma^{(t)}\tilde\gamma^{\star}]&=(1-\eta)\mathbb E[\tilde\gamma^{(t-1)}\tilde\gamma^{\star}]+\eta\mathbb E[f_t(\phi^{(t-1)})f_\star(\phi^\star)]\label{iden2}
\end{align}
where we have introduced the function
\begin{equation}
f_{t,\star}(x,y)\doteq \frac{m_{\nu^{(t-1)},\nu^\star}(x,y)}{\mathbb E[m'_{\nu^{(t-1)},\nu^\star}(x,y)]}-x.
\end{equation}
 
Based on these recursions, we will derive a condition on the parameters of the model for which the assumption of convergence
leads to a contradiction. To this end, we study the asymptotic speed of convergence $\gamma^{(t)}\to \gamma^\star$ which we define as
\begin{equation}
\mu_\gamma\doteq \lim_{t\to \infty}\frac{~\mathcal C_\gamma^\star-\mathcal C_\gamma^{(t)}~}{\mathcal C_\gamma^\star -\mathcal C_\gamma^{(t-1)}}\;.
\end{equation}
The condition $\mu_\gamma\geq 1$ implies that the algorithm no longer converges.  
Note, that a divergence of the algorithm was not observed for the teacher--student matching scenario discussed in \cite{ccakmak2020analysis}. In \ref{Dstabil}, we derive the explicit formula
\begin{equation}
\mu_\gamma=1-\eta \frac{1-\mathbb E[(m'_{\nu^\star}(\gamma^\star,y))^2]{\rm R}'(-\chi^\star)}{1-(\chi^\star)^2{\rm R}'(-\chi^\star)}. \label{stabil}
\end{equation}
Here, from definition of the function ${\rm R}(\omega)$ \eqref{Rtrs} it follows that the term $1-(\chi^\star)^2{\rm R}'(-\chi^\star)$ is always positive. Hence, $\mu_{\gamma }\geq 1$ if and only if 
\begin{equation}
\mathbb E[(m'_{\nu^\star}(\gamma^\star,y))^2]{\rm R}'(-\chi^\star)\geq 1\label{at}.
\end{equation}
Following the arguments in \cite{Adatap,Kab08} we conclude that equation \eqref{at}  coincides
with the stability condition of the {\em replica symmetric} ansatz for the static probabilistic model- known as the {\em de Almeida Thouless} (AT) criterion.
Remarkably, the stability criterion is {\em independent} of the update parameter $\eta$ . This indicates that (at least within our theoretical setting),
a diverging parallel iterated algorithm cannot be made convergent by reverting to a random sequential version.

\section{Simulation results}
\label{Sim}

In the following, we compare our analytical results to numerical simulations of the algorithm for the perceptron model.
 We assume that the teacher model from which data are generated is of the general form \eqref{PercLike} with 
 teacher parameters denoted by $\beta$ and $\sigma$.
 For the student likelihood used in the inference algorithm we restrict ourselves to the simple {\em noise free}
likelihood model
\begin{equation}
p(y\vert \theta )=\Theta(y\theta)
\end{equation}
where $\Theta$ stands for the unit-step function. This is a special case of \eqref{PercLike} corresponding to the limits
$\beta,\sigma\to 0$. We specialise on the following random matrix models for $\matr X$: (i) the entries of $\matr X$ are independent Gaussian with zero mean variance $1/N$; (ii) $\matr X={\matr O}\matr S$ where $\matr S$ is the $N\times P$ projection matrix with $N\geq P$ and $\matr S_{ij}=\delta_{ij}$ for all $i,j$, and $\matr O$ is an $N\times N$ Haar random matrix. The function ${\rm R}(\omega)$ in \eqref{Rtrs} for these models reads as
\begin{equation}
{\rm R}(\omega)=\left\{
\begin{array}{cc} \frac{q-1-\sqrt{(q-1)^2-4\omega}}{2\omega} & \text{model}~ (i)\\
1+\frac{q-1}{\omega}&~\text{model}~ (ii)
\end{array} \right.\;.
\end{equation}
We will first present non asymptotic (finite times) results. 
In order to demonstrate that our analytical approach also applies to non convergent dynamics of the algorithm, 
we consider model parameter settings from the \emph{unstable} region. Specifically, we set $\beta_0=\frac 1 2$ and $\alpha=2$ and $\eta=0.8$. In this case, 
we obtain the following results for two--time correlations over 5 time steps with
\begin{align}
\mathbb E[\phi^{(1:5)}(\phi^{(1:5)})^\top]
&=\tiny{\left[\begin{array}{ccccc}
	6.47 &   11.73 &   17.45 &  23.55 &   29.98\\
	11.73 &  24.03 &   36.60 &   50.15 &  64.46\\
	17.45 & 36.60 &   58.6041 &   80.29 &  103.80 \\
	23.55 & 50.15 & 80.2957 &  114.26 &  146.86 \\
	29.98 &  64.46 &  103.80 & 146.86 &  194.73 
	\end{array}\right] }\nonumber \\
\frac 1 N \matr \phi^{(1:5)}(\matr \phi^{(1:5)})^\top&=
{\tiny\left[ 
	\begin{array}{ccccc}
	6.48 & 11.74&   17.38 &  23.36 &   29.62\\
	11.74 &  24.09&   36.50 &   49.82 &   63.80\\
	17.38 & 36.50 &  58.15 &  79.39 & 102.27\\
	23.36 &49.82  & 79.39 & 112.72 & 144.46\\
	29.62 & 63.80 & 102.27 & 144.46 &  190.87
	\end{array}
	\right]}\nonumber.
\end{align}
We clearly see a strong increase in autocorrelations over time, indicating the divergence of the algorithm.
Here, the simulation result is based on a {\em single instance} of the model with $N=2^{14}$.  
The results were obtained from the random matrix (i). For the  random matrix model (ii) we have similar theory-experiment agreement. 

\begin{figure}[h]
	\epsfig{file=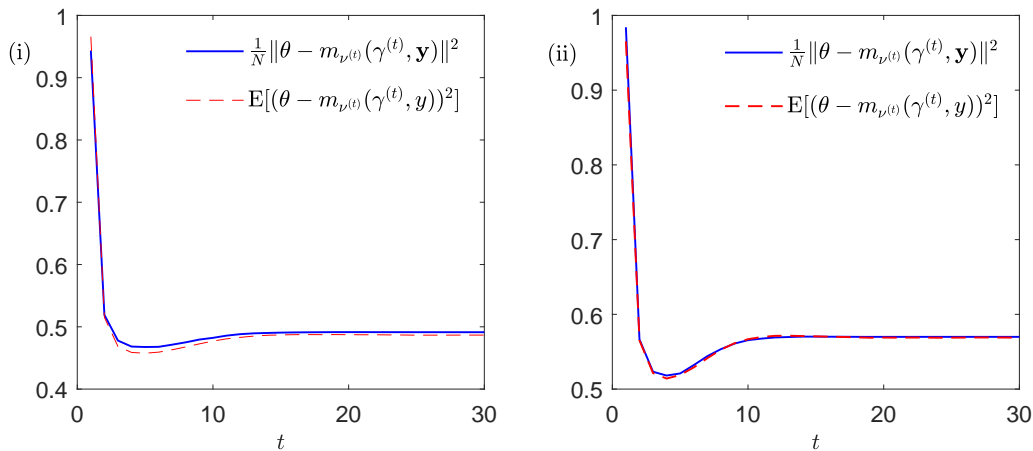,width=1\columnwidth}
	\caption{Predicting of the estimation error for a given iteration time-step $t$.  E.g. the figure with label (i) is for the random matrix model (i). The model parameters are chosen as $\sigma_0^2=10^{-2}$, $\beta_0=0.2$, $\eta=0.5$, $N= 3P/2$ and $P=2^{12}$. }\label{fig1}
\end{figure}

Secondly, we present results on the error of estimating the true teacher parameter $\matr \theta$ at each iteration step in Figure~\ref{fig1}.
The parameters correspond to the region of convergence. In contrast to typical results for cases 
of teacher--student model matching (with optimally chosen variance of initial conditions), the prediction error turns out to be non-monotonic.
Finally, we illustrate the asymptotic speed of convergence predicted by the theory compared to a single simulation of the algorithm.
To show the robustness of our results, we have chosen the parameters yielding large values of static order parameters. Nevertheless, we find a remarkably good prediction of the exponential convergence.

\begin{figure}[h]
	\epsfig{file=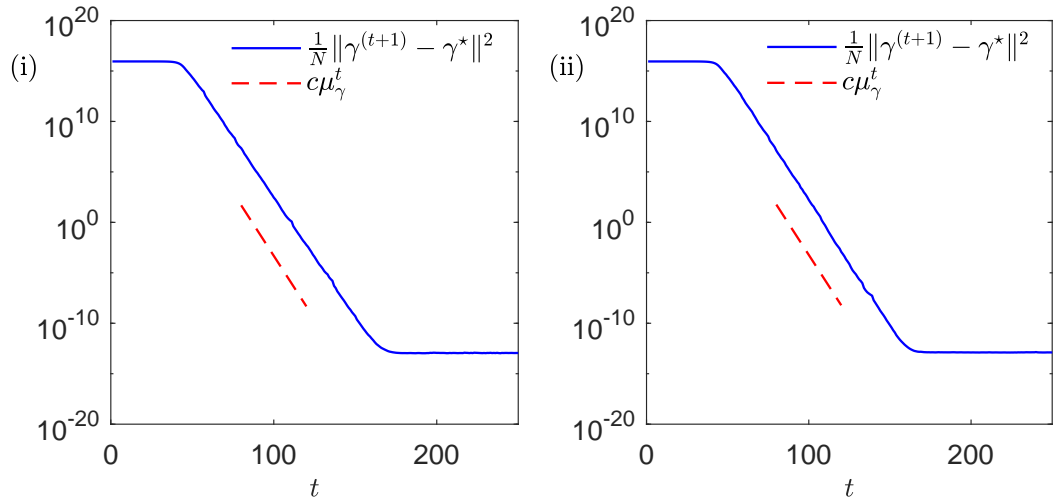,width=1\columnwidth}
	\caption{Asymptotic of the algorithm. The model parameters are chosen as $\sigma_0^2=0.01$, $\beta_0=0.3$, $\eta=0.5$, $N=3P$ and $P=2^{12}$. In this case, we have $\mathcal C_\phi^\star\approx 8\times10^{15}$ and $\hat{\mathcal B}^\star\approx 10^{7}$.} \label{fig2}
\end{figure}

\section{Summary and discussion}
\label{sec:conclusion}

We have analysed the dynamics of a message passing algorithm for inference in large latent Gaussian variable models. Our analysis
is based on a teacher-student scenario together with random matrix assumptions for data. 
We have focused on the problem of student-teacher mismatch and random sequential updates. 
Using a dynamical functional approach
we have decoupled the degrees of freedom and have derived an effective stochastic dynamics for single nodes. 
The absence of memory terms
in the single node dynamics leads to tractable recursions for two--time correlation functions. Comparison between our theory
and simulations on single instances of large systems show excellent agreement. 

We have shown that a teacher-student mismatch opens the possibility of a divergence of the algorithm.~We have identified the range of model parameters for which 
convergence to a fixed point is impossible. Our main result is that the critical set of parameters is identified as the AT line of the static replica symmetric solution. It would be interesting to see if one could prove {\em global convergence} in the stable region. For this one would have to go beyond the local stability analysis presented in this paper and study the full temporal development of a set of coupled order parameters. A possible simplification could be the construction of a Lyapunov function for the single node dynamics.

Since the static AT stability criterion is independent of the update schedule of the dynamics, we were not able to show that a divergent (parallel) algorithm can be made convergent using random sequential updates. One might argue that this negative result could be related to the random matrix distributions used in the modeling of the data. A second possibility is the simplicity of the node update used in our model.
As an alternative one could define random updates of the $\matr{\tilde\gamma}^{(t)}$ in last line of \eqref{svamp1} variables instead. The dynamical mean field analysis of the corresponding model will be given elsewhere.

\section*{Acknowledgment}
This work was supported by the German Research Foundation, Deutsche Forschungsgemeinschaft (DFG), under Grant ``RAMABIM'' with No. OP 45/9-1, by the US National Science Foundation under grant CCF-1910410, and by the Harvard FAS Dean's Competitive Fund for Promising Scholarship.
\appendix
\section{The dynamical functional analysis}\label{Appendixa}
The disorder average in \eqref{disorder} can be computed using the saddle-point method. Specifically, we can follow the steps \cite[Eq. (B.6)--(B.34)]{ccakmak2020analysis}, by essentially replacing all averages over the matrix $\matr A$  by averages over $\matr J$ and read off the result. In our
case, the variables $\matr\theta $, $\matr m^{(k)}$ and $\matr {\hat\psi }^{(k)}$  play the roles of $\matr u$, $\matr \gamma(k)$ and $\matr {\hat \rho}(k)$ in \cite{ccakmak2020analysis}, respectively. Doing so leads to the large $N$ limit approximation of the
the averaged generating functional as
\begin{align}
&\mathbb E[Z_i(\{l(t)\})]\simeq \int {\rm d}\theta{\rm d}y{\rm d} \phi^{(1:T)}\; p_0(y\vert \theta)\mathcal N(\theta\vert 0,q)\mathcal N(\phi^{(1:T)}\vert \theta\mathcal {\hat B},\mathcal C) \prod_{t\leq K}{\rm dP}(p^{(t)}) {\rm d}\psi^{(t)}{\rm d}m^{(t)}\nonumber \\
&\quad \times \delta \left[m^{(t)}- f_t\left(\{\psi^{(l)}, m^{(l)}, p^{(l)}\}_{l=1}^{t};y\right)\right]\delta \left[\psi^{(t)}-\phi^{(t)}-\sum_{t\leq T}\mathcal {\hat G}^{(t,t')}m^{(t)}\right]{\rm e}^{{\rm i}\psi^{(t)}l^{(t)}}.\label{generalresult}
\end{align}
Here, $\hat{\mathcal G}^{(t,l)}$ denotes the $(t,l)$th indexed entries of the $T\times T$ memory matrix $\hat{\mathcal G}$ which is defined in terms of the R-transform 
and its power series expansion as
\begin{equation}
\hat{\mathcal G}={\rm R}(\mathcal G)=\sum_{n=1}^{\infty}c_{n}\mathcal G^{n-1}\label{reseq}.
\end{equation}
Te entries of the $T\times T$ {\em response matrix} $\mathcal G$ are given by
\begin{equation}
\mathcal G^{(t,t')}\doteq\mathbb E\left[\frac{\partial m^{(t)}}{\partial \phi^{(t')}}\right].
\end{equation}
Moreover, the Gaussian process $\{\phi^{(t)}\}$ has the $T\times 1$ mean vector $\theta \mathcal {\hat B}$ and $T\times T$ covariance matrix $\mathcal C$ which are computed by
\begin{align}
\mathcal{\hat B}&=\left(\sum_{n=2}^\infty c_n\sum_{n'=0}^{n-2}(-q)^{n'}\mathcal G^{n-n'-2}\right)\mathcal{B}\label{hat B}\\
\mathcal C&=\sum_{n=2}^\infty c_n\sum_{n'=0}^{n-2}\mathcal G^{n'} \mathcal C_m(\mathcal G^\top)^{n-2-n'}\nonumber \\
&-\sum_{n=3}^\infty c_n\sum_{n'=0}^{n-2}(-q)^{n'}\sum_{l=0}^{n-n'-3}(\mathcal G^\top)^l\mathcal{B}\mathcal{B}^\top\mathcal G^{n-n'-l-3}\label{Cphi}
\end{align}
where we have defined 
\begin{align}
\mathcal C_m^{(t,t')}\doteq \mathbb E[m^{(t)}m^{(t')}]~~\text{and}~~ \mathcal B^{(t)}\doteq -\mathbb E[\theta m^{(t)}].
\end{align}
\subsection{The analysis of sequential dynamics}
By using the property of Dirac-delta function $\delta(\matr y) = \vert\matr X \vert\delta(\matr X\matr y)$ we note that 
\begin{align}
&\delta\left[\matr \phi^{(t)}- \left(\frac 1 {\tau^{(t)}}(\lambda^{(t)}{\bf I}+\matr J)^{-1}-{\bf I}\right)\tilde{\matr \gamma}^{(t)}\right]= \frac {1}{c^{(t)}}\delta\left[\tilde{\matr\gamma}^{(t)}-\tau^{(t)}(\lambda^{(t)}{\bf I}+\matr J)(\matr \phi^{(t)}+\tilde{\matr\gamma}^{(t)})\right]\\
&= \frac 1{c^{(t)}}\int {\rm d}\matr m^{(t)}{\rm d}\matr \psi^{(t)}\delta[\tilde{\matr \gamma}^{(t)}+\lambda^{(t)} \matr m^{(t)}+\matr \psi^{(t)}]\delta[\matr m^{(t)}+\tau^{(t)}(\matr \phi^{(t)}+\tilde{\matr \gamma}^{(t)})]\delta[\matr \psi^{(t)}-\matr J\matr m^{(t)}]
\end{align}
where for short we define the dynamical determinant $c^{(t)}\doteq \left\vert \tau^{(t)}(\lambda^{(t)}{\bf I}+\matr J)\right\vert$ which do not depend on the disorder variables and thereby they solely play the role of appropriate constant terms in the disorder average. Indeed, one can express the dynamics $\matr \phi^{(t)}$ in \eqref{phit} in terms of the system of equations
\begin{subequations}
\begin{align}
\matr m^{(t)}&=-\tau^{(t)}(\matr \phi^{(t)}+\tilde{\matr \gamma}^{(t)})\\
\matr \phi^{(t)}&=\matr J\matr m^{(t)}-\nu^{(t)}\matr m^{(t)}.
\end{align}  
\end{subequations}
\

Consequently, by the general results of the dynamical functional theory \eqref{generalresult}, $\{\phi_i^{(t)}\}_{t=1}^T$ (for an arbitrary component $i$) can be transformed into a Gaussian random sequence by appropriate {\em subtractions}. The 
subtractions define an auxiliary dynamical system which is obtained 
by replacing the variable $\matr \phi^{(t)}$ by 
\begin{equation}
\matr \phi_{aux}^{(t)}= \matr J \matr m^{(t)}-\sum_{l\leq t}\hat{\mathcal G}^{(t,l)}\matr m^{(t)}
\label{auxdyn}
\end{equation}
for $t=1,2,\ldots T$. The entries of the {\em response matrix} $\mathcal G$ read 
\begin{equation}
\mathcal G^{(t,t')}\doteq \mathbb E\left[\frac{\partial m^{(t)}}{\partial \phi_{aux}^{(t')}}\right].
\end{equation}
Moreover, by construction we have 
\begin{equation}
\frac{\partial{\gamma}^{(t-1)}}{\partial {\phi}_{aux}^{(t')}}=\underbrace{p^{(t')}\prod_{l=t'+1}^{t-1}(1-p^{(l)})}_{\doteq p^{(t,t')}}\quad~ t'<t.
\end{equation}
Hence, the response terms read 
\begin{align}
\mathcal G^{(t,t')}&=-\tau^{(t)}\delta_{tt'}-\tau^{(t)}\mathbb E\left[f_{t}(\gamma^{(t-1)};y)p^{(t,t')}\right]\nonumber \\
&=-\tau^{(t)}\delta_{tt'}-\tau^{(t)}{\rm Pr}( p^{(t,t')}=1)\mathbb E\left[f_t'(\gamma^{(t-1)})p^{(t,t')}\vert p^{(t,t')}=1\right]\nonumber \\	&=-\tau^{(t)}\delta_{tt'}-\tau^{(t)}\eta(1-\eta)^{t-1-t'}\mathbb E\left[f_t'(\phi_{aux}^{(t')})\right]\nonumber \\
&=-\tau^{(t)}\delta_{tt'}
\end{align}
where for convenience we have introduced the function 
\begin{equation}
f_t(x,y)\doteq \frac{m_{\nu^{(t-1)}}(x,y)}{\mathbb E[m'_{\nu^{(t-1)}}(x,y)]}-x
\end{equation}
which fulfills the divergence-free property $\mathbb E[f'_t(x,y)]=0$. Thereby, we get
\begin{align}
\matr \phi_{aux}^{(t)}&= \matr J \matr m^{(t)}-\mathcal{\hat G}^{(t,t)}\matr m^{(t)}\\
&=\matr J \matr m^{(t)}-{\rm R}(-\tau^{(t)})\matr m^{(t)}=\matr \phi^{(t)}.
\end{align}
The effective stochastic process (w.r.t. dynamical functional analysis) of $\phi_i^{(1:T)}$ becomes then a Gaussian process as
\begin{equation}
\phi^{(1:K)}\sim \mathcal N(\theta \mathcal {\hat B},\mathcal C)
\end{equation}

We next use the result $\mathcal G^{(t,t')}=-\tau^{(t)}\delta_{tt'}$ to compute the the necessary order parameters  $\mathcal {\hat B}$ and $\mathcal C$ in \eqref{hat B} and \eqref{Cphi}, respectively.
\subsubsection{Computation of  $\mathcal {\hat B}$}
We have
\begin{align}
\mathcal {\hat B}^{(t)}&=\left(\sum_{n=2}^{\infty}c_n\sum_{n'=0}^{n-2}(-q)^{n'}(-\tau^{(t)})^{n-n'-2}\right)\mathcal B^{(t)}\\
&=\frac{\mathcal B^{(t)}}{\tau^{(t)}-q}\sum_{n=2}^{\infty}c_n[(-q)^{n-1}-(-\tau^{(t)})^{n-1}]\\
&=\frac{\mathcal B^{(t)}}{\tau^{(t)}-q}\sum_{n=1}^{\infty}c_n[(-q)^{n-1}-(-\tau^{(t)})^{n-1}]\\
&=\frac{\mathcal B^{(t)}}{\tau^{(t)}-q}({\rm R}(-q)-{\rm R}(-\tau^{(t)}))\\
&=\frac{\mathcal B^{(t)}}{\tau^{(t)}-q}(1/q-{\rm R}(-\tau^{(t)}))
\end{align}
On the other hand, we have
\begin{align}
\mathcal B^{(t)}&=\tau^{(t)}\mathbb E[\theta(\phi^{(t)}+\tilde\gamma^{(t)})]\\
&=\tau^{(t)}(q\mathcal {\hat B}^{(t)}+\mathbb E[\theta\tilde\gamma^{(t)}]).
\end{align}
Combining both result we easily obtain that 
\begin{equation}
\mathcal {\hat B}^{(t)}=\frac{\tau^{(t)}\zeta^{(t)}\mathbb E[\theta\tilde\gamma^{(t)}]}{1-q\tau^{(t)}\zeta^{(t)}} \quad \text{with}\quad \zeta^{(t)}\doteq\frac{{\rm R}(-\tau^{(t)})-1/q}{q-\tau^{(t)}}.
\end{equation}

\subsubsection{Computation of  $\mathcal {C}$}
Recall that
\begin{align}
\mathcal C&=\sum_{n=2}^\infty c_n\sum_{n'=0}^{n-2}\mathcal G^{n'} \mathcal C_m(\mathcal G^\top)^{n-2-n'}\nonumber \\
&\underbrace{-\sum_{n=3}^\infty c_n\sum_{n'=0}^{n-2}(-q)^{n'}\sum_{l=0}^{n-n'-3}(\mathcal G^\top)^l\mathcal{B}\mathcal{B}^\top\mathcal G^{n-n'-l-3}}_{\mathcal D}\label{cphi2}.
\end{align}
We then write 
\begin{align}
\mathcal C^{(t,t')}-\mathcal D^{(t,t')}&=\mathcal C_m^{(t,t')}\sum_{n=2}^\infty c_n \sum_{n'=0}^{n-2}(-\tau^{(t)})^{n'}(-\tau^{(t')})^{n-2-n'}\\
& = \mathcal C_m^{(t,t')}\sum_{n=2}^\infty c_n \frac{(-\tau^{(t')})^{n-1}-(-\tau^{(t)})^{n-1}}{\tau^{(t)}-\tau^{(t')}} \\
&= \mathcal C_m^{(t,t')}\frac{{\rm R}(-\tau^{(t')})-{\rm R}(-\tau^{(t)})}{\tau^{(t)}-\tau^{(t')}}\label{npre}
\end{align}
On the other hand, by construction we have 
\begin{align}
\mathcal C_m^{(t,t')}&=\tau^{(t)}\tau^{(t')}(\mathbb E[\phi^{(t)}\phi^{(t')} ]+ \mathbb E[\phi^{(t)}\tilde\gamma^{(t')}]+\mathbb E[\phi^{(t')}\tilde\gamma^{(t)}]+\mathbb E[\tilde\gamma^{(t)}\tilde\gamma^{(t')}])\\
&=\tau^{(t)}\tau^{(t')}(\mathcal C^{(t,t')}+q\mathcal {\hat B}^{(t)} \mathcal {\hat B}^{(t')} +\mathbb E[\phi^{(t)}\tilde\gamma^{(t')}]+\mathbb E[\phi^{(t')}\tilde\gamma^{(t)}]+\mathbb E[\tilde\gamma^{(t)}\tilde\gamma^{(t')}])\\
&=\tau^{(t)}\tau^{(t')}(\mathcal C^{(t,t')}+q\mathcal {\hat B}^{(t)} \mathcal {\hat B}^{(t')} +\mathcal {\hat B}^{(t)}\mathbb E[\theta\tilde\gamma^{(t')}]+\mathcal {\hat B}^{(t')}\mathbb E[\theta\tilde\gamma^{(t)}]+\mathbb E[\tilde\gamma^{(t)}\tilde\gamma^{(t')}]) \label{result1}
\end{align}
where in the last line we have invoked the results
\begin{align}
\mathbb E[\phi^{(t)}\tilde\gamma^{(t')}]&=\mathbb E[\phi^{(t)}f_{t'}(\gamma^{(t'-1)})]\\&=(1-\eta)\mathbb E[\phi^{(t)}f_{t'}(\gamma^{(t'-2)})]+\eta\mathbb E[\phi^{(t)}f_{t'}(\phi^{(t'-1)})]\\
&=(1-\eta)\mathbb E[\phi^{(t)}f_{t'}(\gamma^{(t'-2)})]+\eta\mathcal {\hat B}^{(t)}\mathbb E[\theta f_{t'}(\phi^{(t'-1)})]\label{cstein}\\
&=\mathcal {\hat B}^{(t)}\mathbb E[\theta\tilde\gamma^{(t')}].
\end{align}
The equation \eqref{cstein} follows from the Stein' lemma.  Then, by invoking \eqref{result1} in \eqref{npre} we get 
\begin{equation}
\mathcal C^{(t,t')}=\frac{\mathcal D^{(t,t')}+\mathcal Q^{(t,t')}\left(q\mathcal {\hat B}^{(t)} \mathcal {\hat B}^{(t')} +\mathcal {\hat B}^{(t)}\mathbb E[\theta\tilde\gamma^{(t')}]+\mathcal {\hat B}^{(t')}\mathbb E[\theta\tilde\gamma^{(t)}]+\mathbb E[\tilde\gamma^{(t)}\tilde\gamma^{(t')}]\right)}{1-\mathcal Q^{(t,t')}}
\end{equation}
We complete the derivation by simplifying $\mathcal D^{(t,t')}$: Firstly, for $\tau^{(t)}\neq\tau^{(t')}$ we have
\begin{align}
\mathcal D^{(t,t')}&=-\mathcal B^{(t)}\mathcal B^{(t')}\sum_{n=3}^\infty c_n\sum_{n'=0}^{n-2}(-q)^{n'}\sum_{l=0}^{n-n'-3}(-\tau^{(t)})^l(-\tau^{(t')})^{n-n'-l-3}\\&=\frac{\mathcal B^{(t)}\mathcal B^{(t')}}{\tau^{(t)}-\tau^{(t')}}\sum_{n=3}^\infty c_n\sum_{n'=0}^{n-2}(-q)^{n'}\{(-\tau^{(t)})^{n-n'-2}-(-\tau^{(t')})^{n-n'-2}\}\\
&=\frac{\mathcal B^{(t)}\mathcal B^{(t')}}{\tau^{(t)}-\tau^{(t')}}\left[\frac{{\rm R}(-\tau^{(t)})-1/q}{q-\tau^{(t)}}-\frac{{\rm R}(-\tau^{(t')})-1/q}{q-\tau^{(t')}}\right]
\end{align}
where we used the fact that 
\begin{align}
\sum_{n=3}^\infty c_n\sum_{n'=0}^{n-2}(-q)^{n'}(-\tau^{(t)})^{n-n'-2}&=\frac{1}{\tau^{(t)}-q}\sum_{n=3}^\infty c_n[(-q)^{n-1}-(-\tau^{(t)})^{n-1}]\\
&=\frac{1}{\tau^{(t)}-q}\sum_{n=1}^\infty c_n[(-q)^{n-1}-(-\tau^{(t)})^{n-1}]-1\\
&=\frac{{\rm R}(-\tau^{(t)})-{\rm R}(-q)}{q-\tau^{(t)}}-c_2\\
&=\frac{{\rm R}(-\tau^{(t)})-1/q}{q-\tau^{(t)}}-c_2
\end{align}
Moreover, in the equal-time case,  we have
\begin{equation}
\mathcal D^{(t,t)}=\frac{\mathcal B^{(t)}\mathcal B^{(t)}}{q-\tau^{(t)}}\left[\frac{{\rm R}(-\tau^{(t)})-1/q}{q-\tau^{(t)}}-{\rm R}'(-\tau^{(t)})\right].
\end{equation}
\section{Derivation of \eqref{stabil}}\label{Dstabil}
For convenience, we define the single time deviations $\Delta_{\gamma,\phi}^{(t)}\doteq \mathcal C_\gamma^\star-\mathcal C_{\gamma,\phi}^{(t)}$. Furthermore, from \eqref{Cgamma} we write the recursion
\begin{equation}
\Delta_{\gamma}^{(t)}=(1-\eta)\Delta_{\gamma}^{(t-1)}+\eta \Delta_{\phi}^{(t)}.\label{newdev}
\end{equation}
Moreover, we introduce the rate $\mu_\phi\doteq \lim_{t\to \infty}\frac{\Delta_\phi^{(t)}}{\Delta_\phi^{(t-1)}}$. From \eqref{newdev} it follows that $\mu_\gamma=\mu_\phi$. We next compute $\mu_\phi$.  To this end, from \eqref{Cphi1} and \eqref{Cphi2} we firstly write $\Delta_{\phi}^{(t)}$ in the form 
\begin{align}
\Delta_\phi^{(t)}=\mathcal C_\gamma^\star-c^{(t)}\hat{\mathcal B}^\star\hat{\mathcal B}^{(t)}-\frac{\mathcal Q^{(t)}}{1-\mathcal Q^{(t)}}\mathbb E[\tilde\gamma^{(t)}\tilde\gamma^{\star}]
\end{align}
for an appropriately computed constant sequence $c^{(t)}$ (which does not depend on $\Delta_\phi^{(t-1)}$). Then, from \eqref{iden2} we further write 
\begin{align}
\Delta_\phi^{(t)}&=\mathcal C_\gamma^\star-c^{(t)}\hat{\mathcal B}^\star\hat{\mathcal B}^{(t)}-(1-\eta)\frac{\mathcal Q^{(t)}(1-\mathcal Q^{(t-1)})}{(1-\mathcal Q^{(t)})\mathcal Q^{(t-1)}}(\mathcal C_\gamma^\star-c^{(t-1)}\hat{\mathcal B}^\star\hat{\mathcal B}^{(t-1)})\nonumber \\
&+(1-\eta)\frac{\mathcal Q^{(t)}(1-\mathcal Q^{(t-1)})}{(1-\mathcal Q^{(t)})\mathcal Q^{(t-1)}}\Delta_\phi^{(t-1)}-\eta\frac{\mathcal Q^{(t)}}{1-\mathcal Q^{(t)}}\mathbb E[f'_t(\phi^{(t-1)})f'_\star(\phi^\star)].
\end{align}
Thereby, we get the derivative 
\begin{equation}
g_{t}(\Delta_\phi^{(t-1)})\doteq\frac{\partial\Delta_\phi^{(t)}}{\partial\Delta_\phi^{(t-1)}}= (1-\eta)\frac{\mathcal Q^{(t)}(1-\mathcal Q^{(t-1)})}{(1-\mathcal Q^{(t)})\mathcal Q^{(t-1)}}+ \eta\frac{\mathcal Q^{(t)}}{1-\mathcal Q^{(t)}}\mathbb E[f'_t(\phi^{(t-1)})f'_\star(\phi^\star)].
\end{equation}
We then obtain the rate as
\begin{align}
\mu_\phi&=\lim_{t\to \infty}g_{t}(0)= (1-\eta)+\eta \frac{(\chi^\star)^2{\rm R}'(-\chi^\star)}{1-(\chi^\star)^2{\rm R}'(-\chi^\star)}\mathbb E[(f'_\star(\phi^\star))^2]\\
&= (1-\eta)+\eta \frac{{\rm R}'(-\chi^\star)}{1-(\chi^\star)^2{\rm R}'(-\chi^\star)}(\mathbb E[(m_{\nu^\star}'(\phi^\star))^2]-(\chi^\star)^2)\\
&=1-\eta \frac{1-\mathbb E[(m_{\nu^\star}'(\phi^\star))^2]{\rm R}'(-\chi^\star)}{1-(\chi^\star)^2{\rm R}'(-\chi^\star)}.
\end{align} 
\bibliographystyle{iopart-num}
\bibliography{mybib}
\end{document}